\pdfoutput=1

\documentclass[11pt]{article}

\usepackage[preprint]{acl}

\usepackage{times}
\usepackage{latexsym}

\usepackage[T1]{fontenc}

\usepackage[utf8]{inputenc}

\usepackage{microtype}
\usepackage{inconsolata}

\usepackage{hyperref}
\usepackage{url}

\usepackage{booktabs}
\usepackage{nicefrac}
\usepackage{xcolor}
\usepackage{graphicx}
\usepackage{wrapfig}
\usepackage{subcaption}
\usepackage{dirtytalk}
\usepackage{textcomp}
\usepackage{listings}
\usepackage{pythonhighlight}
\usepackage{amsmath}
\usepackage{amssymb}
\usepackage{amsthm}
\usepackage{amsfonts}
\usepackage{cleveref}
\usepackage{makecell}
\usepackage{multicol}
\usepackage{multirow}
\usepackage{bbding}
\usepackage{colortbl}
\usepackage{pifont}
\usepackage{mathrsfs}
\usepackage{bbm}
\usepackage{bm}
\usepackage{enumitem}
\usepackage{xspace}
\usepackage{dsfont}
\usepackage{soul}
\usepackage{tikz}
\usepackage{centernot}
\usepackage{tcolorbox}

\title{SWI: Speaking with Intent in Large Language Models}

\author{Yuwei Yin$^{1}$, EunJeong Hwang$^{1,2}$, Giuseppe Carenini$^{1}$ \\ 
$^{1}$University of British Columbia; $^{2}$Vector Institute for AI \\
\texttt{\{yuweiyin,ejhwang,carenini\}@cs.ubc.ca}
}

\definecolor{darkblue}{rgb}{0, 0, 0.5}
\hypersetup{colorlinks=true, citecolor=darkblue, linkcolor=darkblue, urlcolor=darkblue}
\definecolor{royalblue}{rgb}{0.255,0.412,0.882}
\definecolor{deepblue}{rgb}{0,0,0.5}
\definecolor{deepred}{rgb}{0.5,0,0}
\definecolor{deepgreen}{rgb}{0,0.5,0}
\definecolor{halfgray}{gray}{0.5}
\definecolor{blue_}{rgb}{0,0,1}
\definecolor{green_}{rgb}{0,0.5,0}
\definecolor{beaublue}{rgb}{0.74, 0.83, 0.9}
\definecolor{beige}{rgb}{0.96, 0.96, 0.86}
\definecolor{bisque}{rgb}{1.0, 0.89, 0.77}
\definecolor{linen}{rgb}{0.98, 0.94, 0.9}
\definecolor{lightyellow}{rgb}{1.0, 1.0, 0.88}
\definecolor{lightgreen}{rgb}{0.56, 0.93, 0.56}
\definecolor{lightblue}{rgb}{0.68, 0.84, 0.90}
\definecolor{teagreen}{rgb}{0.82, 0.94, 0.75}
\definecolor{turquoisegreen}{rgb}{0.63, 0.84, 0.71}
\definecolor{lawngreen}{rgb}{0.49, 0.99, 0.0}
\definecolor{mintgreen}{rgb}{0.6, 1.0, 0.6}
\definecolor{darkseagreen}{rgb}{0.56, 0.74, 0.56}
\definecolor{swi_yellow}{rgb}{1.0, 1.0, 0.848}
\definecolor{swi_green}{rgb}{0.95, 1.0, 0.94}
\definecolor{swi_green_dark}{rgb}{0.88, 1.0, 0.88}
\definecolor{lightgray}{gray}{0.9}

\begin{document}

\maketitle

\begin{abstract}
Intent, typically clearly formulated and planned, functions as a cognitive framework for communication and problem-solving.
This paper introduces the concept of \textbf{Speaking with Intent (SWI)} in large language models (LLMs), where the explicitly generated intent encapsulates the model's underlying intention and provides high-level planning to guide subsequent analysis and action.
By emulating deliberate and purposeful thoughts in the human mind, SWI is hypothesized to enhance the reasoning capabilities and generation quality of LLMs.
Extensive experiments on text summarization, multi-task question answering, and mathematical reasoning benchmarks consistently demonstrate the effectiveness and generalizability of Speaking with Intent over direct generation without explicit intent.
Further analysis corroborates the generalizability of SWI under different experimental settings.
Moreover, human evaluations verify the coherence, effectiveness, and interpretability of the intent produced by SWI.
The promising results in enhancing LLMs with explicit intents pave a new avenue for boosting LLMs' generation and reasoning abilities with cognitive notions.\footnote{Source code: \url{https://github.com/YuweiYin/SWI}}
\end{abstract}

\section{Introduction}
\label{sec:introduction}

\begin{figure}[t!]
  \centering
  \begin{subfigure}[b]{0.51\linewidth}
    \includegraphics[width=\linewidth]{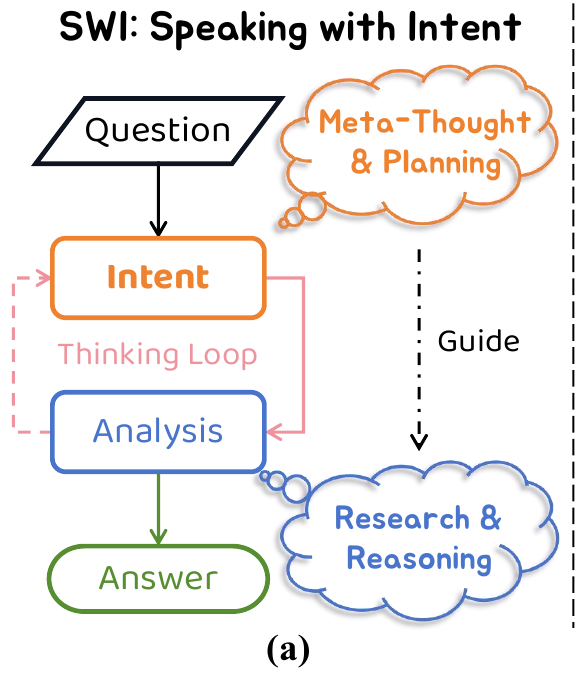}
    \label{fig:swi_overview_1}
  \end{subfigure}
  \vspace{-10pt}
  \begin{subfigure}[b]{0.47\linewidth}
    \includegraphics[width=\linewidth]{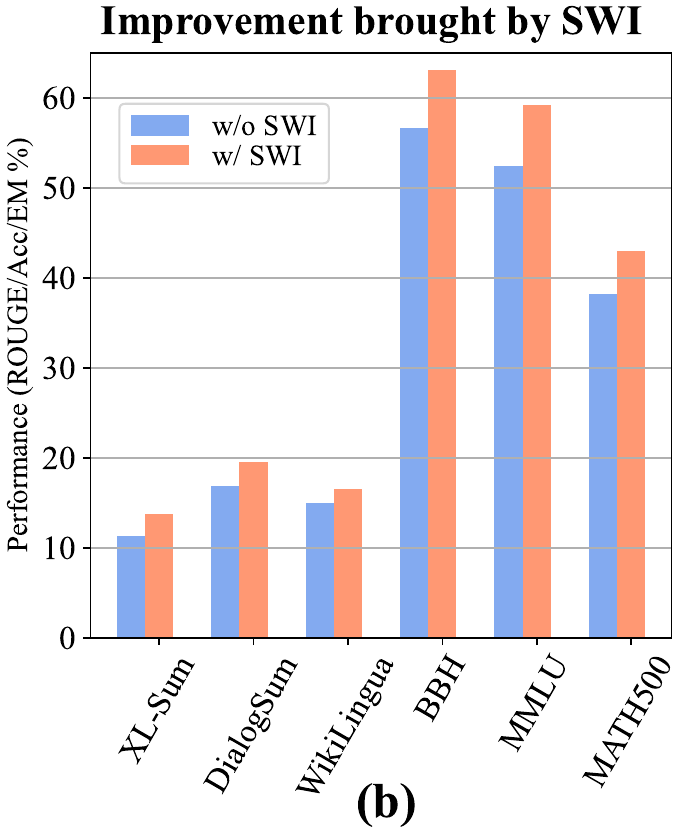}
    \label{fig:swi_overview_2}
  \end{subfigure}
  \vspace{-20pt}
  \caption{\textbf{SWI Overview.} \textbf{(a)} The intent, functioning as meta-thought and planning, guides the analysis with reasoning to answer the question. \textbf{(b)} The performance improvements brought by SWI on various tasks.}
  \label{fig:swi_overview}
  \vspace{-15pt}
\end{figure}

Intent, the goal-oriented intention in our mind~\citep{adams1986intention,mele1989intention,mele1994intentional}, serves as a critical component in communication and a guiding framework for problem-solving.
As illustrated in Figure~\ref{fig:swi_overview}(a), human thinking~\citep{kahneman2011thinking} typically follows a structured loop where intent---a mental state or proactive commitment to perform a specific action or produce a particular outcome---directs problem analysis and logical reasoning, therefore also facilitating communication and interaction.
Hence, we hypothesize that enabling AI systems to speak with their intent explicitly can replicate this meta-cognitive process, thereby improving their generation quality and reasoning ability.

In recent years, large language models (LLMs)~\citep{zhao2023llm_survey,min2023llm_survey,minaee2024llm_survey} have revolutionized Natural Language Processing (NLP) with their excellent generative capabilities~\citep{openai2024gpt4o,anthropic2024claude3,google2024gemini1_5}.
Enhancing LLMs on various language understanding and logical reasoning tasks~\citep{hendrycks2021mmlu,suzgun2023bbh,lightman2024math500,vendrow2025gsm8k_platinum} is vital for their ongoing development~\citep{huang2023reasoning_survey,qiao2023reasoning_survey,patil2025advancing_reasoning}.

This work introduces Speaking with Intent (SWI), requiring LLMs to articulate their own intent as a planning mechanism during generation, which makes LLM intentionality more explicit, in a way reminiscent of the long tradition of intent-driven generation in classic NLG~\citep{grosz1986attention,mann1988rhetorical,moore1993planning,reiter2000nlg}. 
In essence, we hypothesize that, due to the autoregressive nature of LLMs~\citep{openai2019gpt2} and the attention mechanism~\citep{vaswani2017transformer}, explicitly stated intent provides high-level guidance for subsequent analysis and reasoning.
For example, when applying SWI to summarization tasks, each analytical step in summarizing an article is effectively guided by a preceding intent statement, which is a piece of free-form text generated by instruction-following LLMs~\citep{ouyang2022rlhf,rafailov2023dpo,meta2024llama3} instead of a predefined class as in traditional intent modeling~\citep{weld2022survey_intent_detection}.

In this work, we verify the proposed hypothesis by comprehensively evaluating the effectiveness and generalizability of the proposed SWI method.
Specifically, the experimental results across three diverse task categories (i.e., text summarization, multi-task question answering, and mathematical reasoning) demonstrate that speaking with intent in LLMs consistently outperforms directly generating responses without explicit intent.
In summarization tasks, SWI produces summaries that are more accurate, concise, and factually reliable, with fewer hallucinations~\citep{ji2023hallucination_survey,li2024factuality_hallucination} in the output.
In math reasoning tasks, SWI surpasses the LLM reasoning method Chain-of-Thought (CoT)~\citep{kojima2022cot_think_step_by_step} and LLM planning method Plan-and-Solve (PS)~\citep{wang2023plan}, and SWI can work synergistically with these methods to further improve them.
Additionally, we perform human evaluations to assess the coherence, effectiveness, and interpretability of the intent generated by our SWI method.
Evaluators largely agree on the quality of the generated intent, particularly for mathematical reasoning tasks.
The evaluation results confirm that SWI enhances task performance and output explainability.

The key contributions of this work are as follows:
\ding{202} We introduce Speaking with Intent in LLMs, where the generated intent effectively guides problem analysis, logical reasoning, and text generation, boosting performance across various benchmarks.
\ding{203} Extensive experiments and analyses across diverse task types and multiple datasets, including text summarization, multi-task QA, and mathematical reasoning, demonstrate the consistent effectiveness and generalizability of SWI.
\ding{204} Human evaluations validate the coherence, effectiveness, and interpretability of the intent generated by SWI, with our evaluation practice providing standards for assessing freely generated intents.

\section{Speaking with Intent}
\label{sec:method}

This section presents the problem-solving workflow of LLMs and introduces Speaking with Intent (SWI), enabling LLMs to explicitly articulate their intent during response generation.

\subsection{Problem-solving Workflow using LLMs}
\label{subsec:method_workflow}

Let $\mathcal{D} = \{\mathcal{X}, \mathcal{Y}\}$ be a dataset, where $\mathcal{X} = \{X_1, X_2, \dots, X_n\}$ is the input information (questions), $\mathcal{Y} = \{y_1, y_2, \dots, y_n\}$ is the corresponding references (correct answers), and $n$ is the number of instances in $\mathcal{D}$.
For text summarization datasets, $X_i$ is the source article, and $y_i$ is one of the reference summaries.
For multiple-choice QA datasets, $X_i$ contains the question and options, and $y_i$ is the answer label such as (A)/(B)/(C) or Yes/No.
For mathematical reasoning datasets, $X_i$ is the math problem, and $y_i$ is the correct answer (usually an integer number).

In this work, we employ instruction-following LLMs (aka Chat LLMs) $\mathcal{M}$ for experiments and apply the chat template with the system prompt $P_s$ and user prompt $P_u$.
The system prompt specifies the general behavior of the model (assistant), and the user prompt poses questions to the model.
Therefore, the generated output $\hat{y}_i$ is obtained by
\begin{align}\label{equ:llm_gen}
    \hat{y}_i = \mathcal{M}(P_s, P_u, X_i; \Theta, \zeta),
\end{align}
where $P_s$ is the system prompt, $P_u$ is the user prompt, and $X_i$ is the task input.
These string objects are concatenated using line breaks (``$\backslash$n'') as the delimiter.
With parameters $\Theta$ and hyper-parameters $\zeta$, the LLM $\mathcal{M}$ generates new tokens one by one until reaching the generation limit or generating the ``end-of-text'' special token provided by the tokenizer.

\subsection{LLM Speaking with Intent}
\label{subsec:method_swi}

SWI is a novel method that brings a significant cognitive concept (i.e., intent) into LLM generation.
SWI is implemented in a simple and reproducible approach, i.e., we require LLMs to speak with intent (SWI) by presenting detailed instructions in the system prompts $P_s$ and restating the SWI requirement in the user prompt $P_u$.
Table~\ref{tab:swi_instruction} presents the SWI instructions for QA and math tasks. For summarization tasks, the fourth requirement in the system prompt is \textit{At last, clearly and concisely give your final summary starting with "Final Summary:"} and the user prompt becomes \textit{Speak with intent and summarize the following article.} \texttt{\{\{article\}\}}.

\begin{figure}[t!]
  \centering
  \includegraphics[width=0.98\linewidth]{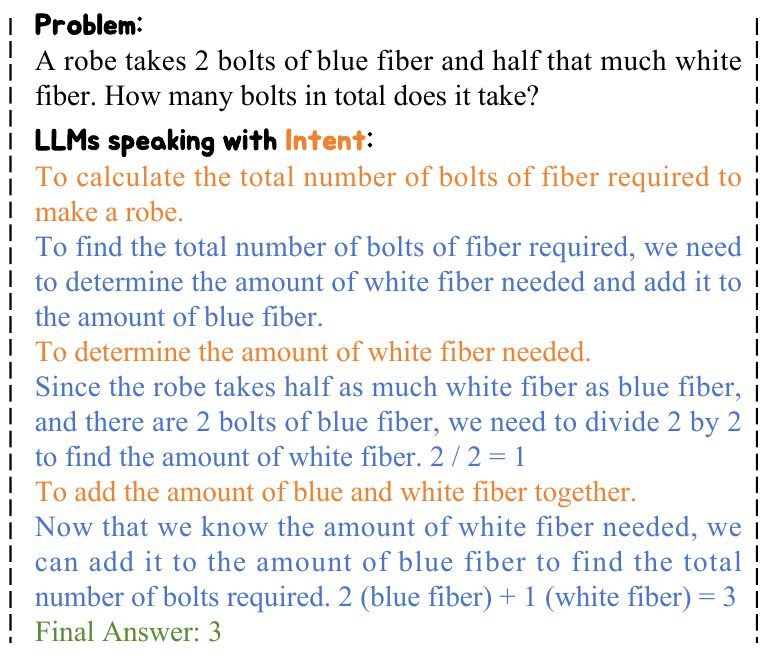}
  \caption{\textbf{An example of LLMs speaking with intent} for mathematical reasoning tasks.}
  \label{fig:swi_showcase}
\end{figure}

As showcased in Figure~\ref{fig:swi_showcase} (math) and Figure~\ref{fig:case_study} (summarization), LLMs that speak with intent articulate their \textcolor{orange}{intents} during the thinking, reasoning and communication process and then provide the final answer based on the analysis.

\begin{table}[t!]
    \centering
    \scalebox{0.98}{
    \begin{tcolorbox}[colback=black!5!white,colframe=black!75!black]
    \normalsize{\textbf{System Prompt}:} \small{You are a helpful assistant who speaks with intent. \\ During generation, follow all the requirements below: \\
    1. Always explicitly state your own intent before speaking each sentence. \\
    2. Each intent statement should explain the sentence that follows. \\
    3. Your intent must start with the "<INTENT>" tag and end with the "</INTENT>" tag. The content within the intent tags must begin with "To" followed by a verb, such as "To accomplish a task." \\
    4. At last, clearly and concisely give your final answer starting with "Final Answer:"} \vspace{5pt} \\
    \normalsize{\textbf{User Prompt}:} \small{Speak with intent and answer the following question. \\ \texttt{\{\{question\}\}} }
    \end{tcolorbox}
    }
    \caption{\textbf{SWI instructions for QA and math tasks.}}
    \label{tab:swi_instruction}
    \vspace{-10pt}
\end{table}

\subsection{Result Evaluation}
\label{subsec:method_evaluation}

We extract the final answer (denoted as $\tilde{y}_i$) from the model outputs $\hat{y}_i$ and compute the overall performance of $\mathcal{M}$ on the dataset $\mathcal{D}$ by
\begin{align}\label{equ:evaluation}
    s = \frac{1}{n} \sum_{i=1}^{n} \mathcal{S}(y_i, \tilde{y}_i),
\end{align} where the score function $\mathcal{S}(\cdot, \cdot)$ returns a value in the range of $[0, 1]$. Different tasks adopt different score functions to evaluate the model performance.

To evaluate the quality of summaries, we apply the standard ROUGE~\citep{lin2004rouge} as the automatic evaluation metric $\mathcal{S}$ and complement it with a more sophisticated fact-checking analysis as described in Section~\ref{subsec:exp_sum}.
For multiple-choice QA tasks, we adopt the Option Selection metric introduced by \citet{yin2025arr}, which evaluates the LLM perplexity of different option concatenations and selects the one with lowest perplexity as the model's choice.
For mathematical reasoning tasks, we first extract numbers in $\tilde{y}_i$ and apply text normalization to both $\tilde{y}_i$ and the reference $y_i$, and then conduct exact match to check if the generated answer $\tilde{y}_i$ is correct.

\section{Experimental Setup}
\label{sec:exp_setup}

This section presents the experimental setup, including Tasks and Datasets (\S~\ref{subsec:exp_setup_data}), Model Settings (\S~\ref{subsec:exp_setup_model}), and Baseline Settings (\S~\ref{subsec:exp_setup_baseline}).

\subsection{Tasks and Datasets}
\label{subsec:exp_setup_data}

The proposed SWI method aims to enhance the generation quality and reasoning ability of LLMs.
To comprehensively study the effectiveness and generalizability of SWI, we conduct extensive experiments on various challenging benchmarks of three diverse task types: text summarization (Sum), multi-task question answering (QA), and mathematical reasoning (Math).
The dataset statistics are presented in Table~\ref{tab:dataset_stat}.

\paragraph{Text Summarization (Sum).}
We hypothesize that Speaking with Intent benefits natural language generation tasks like summarization, where the generated intent can guide the model in summarizing the source article point by point in an orderly fashion.
Hence, we test the effect of SWI on the following text summarization datasets covering different genres: CNN/DailyMail (CDM)~\citep{hermann2015cnn_dailymail_1,see2017cnn_dailymail_2}, Extreme summarization (XSum)~\citep{narayan2018xsum}, XL-Sum~\citep{hasan2021xl_sum}, DialogSum~\citep{chen2021dialogsum}, and WikiLingua~\citep{ladhak2020wikilingua}.

\begin{table}[t!]
    \centering
    \scalebox{0.8}{
    \begin{tabular}{clcc}
    \toprule
    \midrule
    \textbf{Task} & \textbf{Dataset} & \textbf{Split} & \textbf{Size} \\
    \midrule
    \multirow{5}{*}{\makecell[c]{Sum}} & CDM & Test & 11,490 \\
    & XSum & Test & 11,334 \\
    & XL-Sum & Test & 11,535 \\
    & DialogSum & Test & 1,500 \\
    & WikiLingua & Test & 3,000 \\
    \midrule
    \multirow{3}{*}{\makecell[c]{QA}} & BBH & Test & 5,511 \\
    & MMLU & Test & 13,842 \\
    & MMLU-Pro & Test & 12,032 \\
    \midrule
    \multirow{3}{*}{\makecell[c]{Math}} & GSM8K & Test & 1,319 \\
    & GSM8K-P & Test & 1,209 \\
    & MATH500 & Test & 500 \\
    \midrule
    \bottomrule
    \end{tabular}
    }
    \caption{\textbf{Dataset Statistics.}}
    \label{tab:dataset_stat}
    \vspace{-10pt}
\end{table}

\paragraph{Multi-task Question Answering (QA).}
Our SWI method is also evaluated on various multi-task question answering datasets, including BIG-Bench Hard (BBH)~\citep{suzgun2023bbh}, MMLU~\citep{hendrycks2021mmlu}, and MMLU-Pro~\citep{wang2024mmlu_pro}. They are all reasoning-intensive benchmarks designed as multiple-choice QA tasks, where the model is asked to select the most appropriate one from the given options to answer the question.
Here, the hypothesis is that generating its intent explicitly as text improves the system's question analysis abilities.

\paragraph{Mathematical Reasoning (Math).}
Beyond multiple-choice QA tasks, we also consider mathematical reasoning benchmarks, where the model is asked to solve the given math problem and present the final answer (numerical values).
We consider representative and high-quality math benchmarks, including Grade School Math 8K (GSM8K)~\citep{cobbe2021gsm8k}, GSM8K-Platinum~\citep{vendrow2025gsm8k_platinum}, and MATH500~\citep{lightman2024math500}.
Again, our hypothesis is that speaking with explicit intent improves the model's reasoning abilities.

\subsection{Model Settings}
\label{subsec:exp_setup_model}

\paragraph{Language Models.}
By default, we employ LLaMA3-8B-Instruct~\citep{meta2024llama3} as the language model $\mathcal{M}$ for generation and evaluation. It is an open-weights, instruction-following, and Transformer-based~\citep{vaswani2017transformer} LLM with $8$ billion model parameters. We load the model checkpoint and tokenizer provided by Hugging Face Transformers~\citep{wolf2020transformers}.
To further assess the generality of SWI, we also evaluate the efficacy of SWI with different LLMs in \S~\ref{subsec:more_llms}.
In the fact-checking evaluation for summaries (\S~\ref{subsec:exp_sum_fact_check}), we adopt GPT-4o-mini~\citep{openai2024gpt4o} to decompose the generated summary and reference summary into two sets of atomic facts.

\paragraph{Generation Configurations.}
Each experiment session was conducted on a single NVIDIA H100 GPU, and all the models were loaded in a half-precision mode (\texttt{float16}).
The input sequence is not truncated to avoid losing context information, while we set the maximum number of newly generated tokens to $4096$ during generation.

\paragraph{Reproducibility Statement.}
To guarantee reproducibility, we fixed the seeds to 42 for all random modules, set the LLM generation temperature to 0 for deterministic generation without sampling, and ran all experiments three times, obtaining reproducible generation outputs and evaluation scores.
The source code is available on GitHub.

\subsection{Baseline Settings}
\label{subsec:exp_setup_baseline}

The main comparison is LLM generation with intent or without intent, and the effectiveness of SWI is verified if the former outperforms the latter.

In \S~\ref{subsec:synergy_with_cot_ps}, we also investigate the synergy between SWI and existing LLM reasoning \& planning methods, i.e., Chain-of-Thought (CoT)~\citep{kojima2022cot_think_step_by_step} and Plan-and-Solve (PS)~\citep{wang2023plan}.
CoT aims to elicit LLM reasoning using the answer-trigger prompt $\Phi_i^{\text{CoT}}$ as ``\textit{Let's think step by step}'', while PS applies the following prompt $\Phi_i^{\text{PS}}$ to construct plans before problem-solving: ``\textit{Let's first understand the problem and devise a plan to solve the problem. Then, let's carry out the plan and solve the problem step by step.}''
With such answer-trigger prompts $\Phi_i$, the generation process (Eq.~\ref{equ:llm_gen}) is given by 
\begin{align}\label{equ:llm_gen_phi}
    \hat{y}_i = \mathcal{M}(P_s, P_u, X_i, \Phi_i; \Theta, \zeta),
\end{align}
where $P_s$ is the system prompt, $P_u$ is the user prompt, and $X_i$ is the task input. $\Theta$ and $\zeta$ are the parameters and hyper-parameters of the LLM $\mathcal{M}$, respectively.

\section{Main Results}
\label{sec:exp_main_results}

This section presents the experimental results to verify the effectiveness of SWI on diverse generation and reasoning tasks.

\begin{table*}[t!]
    \centering
    \scalebox{0.65}{
    \begin{tabular}{ccccccccccccc}
    \toprule
    \midrule
    & \multirow{2}{*}{\textbf{Method}} & \multicolumn{5}{c}{\textbf{Text Summarization} (Average ROUGE-1/2/L/Lsum \%)} & \multicolumn{3}{c}{\textbf{Multi-task QA} (Accuracy \%)} & \multicolumn{3}{c}{\textbf{Math Reasoning} (Accuracy \%)} \\
    \cmidrule(lr){3-7} \cmidrule(lr){8-10} \cmidrule(lr){11-13}
    & & \textbf{CDM} & \textbf{XSum} & \textbf{XL-Sum} & \textbf{DialogSum} & \textbf{WikiLingua} & \textbf{BBH} & \textbf{MMLU} & \textbf{MMLU-Pro} & \textbf{GSM8K} & \textbf{GSM8K-P} & \textbf{MATH500} \\
    \midrule
    \ding{192} & \multirow{1}{*}{w/o SWI} & 23.38 & 11.90 & 11.29 & 16.92 & 15.01 & 56.65 & 52.40 & 39.27 & 79.08 & 81.89 & 38.20 \\
    \midrule
    \rowcolor{swi_yellow}\ding{193} & \multirow{1}{*}{\textbf{SWI}} & \textbf{23.91} & \textbf{13.90} & \textbf{13.80} & \textbf{19.57} & \textbf{16.53} & \textbf{63.11} & \textbf{59.22} & \textbf{43.72} & \textbf{80.06} & \textbf{82.88} & \textbf{43.00} \\
    \midrule
    \bottomrule
    \end{tabular}
    }
    \caption{\textbf{Main results} on text summarization, multi-task QA, and mathematical reasoning tasks.}
    \label{tab:exp_main_results}
    \vspace{-10pt}
\end{table*}

\subsection{Text Summarization}
\label{subsec:exp_sum}

First, we demonstrate that SWI benefits natural language generation tasks like summarization by more explicitly analyzing the source document point by point and better planning the generation of the final summary.

We evaluate the quality of summaries using the ROUGE score~\citep{lin2004rouge}, which counts the overlaps of the generated summaries and reference summaries.
Specifically, we average the ROUGE-1 (unigrams), ROUGE-2 (bigrams), ROUGE-L (longest common subsequences), and ROUGE-LSum (sentence-level ROUGE-L) scores as the final ROUGE score.
As shown in Table~\ref{tab:exp_main_results}, our SWI method consistently surpasses the direct generation baseline (``w/o SWI''), confirming its effectiveness in enhancing the quality of text summaries.

\subsection{Multi-task Question Answering}
\label{subsec:exp_qa}

Beyond text understanding and generation, we also consider multi-task QA tasks.
We test the effect of SWI on three large-scale and challenging benchmarks designed as multiple-choice QA tasks, where the model is asked to select the most appropriate one from the given options to answer the question.
Table~\ref{tab:exp_main_results} shows that our SWI method consistently improves the direct generation baseline by a large margin. The results demonstrate the efficacy of SWI in reasoning-intensive QA tasks.

\subsection{Mathematical Reasoning}
\label{subsec:exp_math}

Additionally, we explore the efficacy of SWI on three high-quality math benchmarks.
Unlike multiple-choice QA in \S~\ref{subsec:exp_qa}, where the model picks an option from the given list, math tasks require LLMs to generate numerical values as the answer.
As shown in Table~\ref{tab:exp_main_results}, SWI consistently improves the model performance over direct generation, showing its effectiveness in enhancing LLM on problem analysis and mathematical reasoning.

\section{Analysis}
\label{sec:analysis}

\begin{table}[t!]
    \centering
    \scalebox{0.7}{
    \begin{tabular}{cccccc}
    \toprule
    \midrule
    \textbf{Dataset} & & \textbf{Method} & \textbf{Precision} & \textbf{Recall} & \textbf{F1} \\
    \midrule
    \multirow{2}{*}{CDM} & \ding{192} & w/o SWI & 26.06 & \underline{76.28} & 36.37 \\
    & \ding{193} & \cellcolor{swi_yellow}\textbf{SWI} & \underline{34.22} & 55.89 & \cellcolor{swi_yellow}\textbf{37.79} \\
    \midrule
    \multirow{2}{*}{XSum} & \ding{194} & w/o SWI & 11.06 & \underline{48.38} & 15.15 \\
    & \ding{195} & \cellcolor{swi_yellow}\textbf{SWI} & \underline{14.77 }& 37.30 & \cellcolor{swi_yellow}\textbf{16.29} \\
    \midrule
    \multirow{2}{*}{XL-Sum} & \ding{196} & w/o SWI & 8.96 & \underline{61.88} & 13.79 \\
    & \ding{197} & \cellcolor{swi_yellow}\textbf{SWI} & \underline{12.96} & 46.72 & \cellcolor{swi_yellow}\textbf{16.51} \\
    \midrule
    \multirow{2}{*}{DialogSum} & \ding{198} & w/o SWI & 23.99 & \underline{57.08} & 29.55 \\
    & \ding{199} & \cellcolor{swi_yellow}\textbf{SWI} & \underline{34.92} & 45.19 & \cellcolor{swi_yellow}\textbf{31.20} \\
    \midrule
    \multirow{2}{*}{WikiLingua} & \ding{200} & w/o SWI & 23.33 & \underline{65.55} & 30.63 \\
    & \ding{201} & \cellcolor{swi_yellow}\textbf{SWI} & \underline{32.40} & 54.98 & \cellcolor{swi_yellow}\textbf{35.78} \\
    \midrule
    \bottomrule
    \end{tabular}
    }
    \caption{\textbf{Fact Checking Evaluation of Summaries.} We compare the atomic facts drawn from the LLM-generated summaries and  the golden references, and compute recall, precision, and F1 scores (\%).}
    \label{tab:exp_sum_fact_check}
    \vspace{-5pt}
\end{table}

\subsection{Fact Checking of Summaries}
\label{subsec:exp_sum_fact_check}

LLMs frequently generate hallucinated content~\citep{ji2023hallucination_survey,li2024factuality_hallucination}, which can not be detected by lexical metrics like ROUGE.
To assess this issue, we adopt a more semantically sophisticated fact-checking metric~\citep{hwang2025bottlehumor}, which quantifies factual consistency by calibrating the extent of fabricated statements (low precision) and omitted factual information (low recall).
Specifically, we use GPT-4o-mini~\citep{openai2024gpt4o} to decompose both generated and reference summaries into atomic fact sets and measure their coverage to quantify factual consistency.

We evaluate $100$ random samples from each summarization dataset using this fact-checking metric, with results presented in Table~\ref{tab:exp_sum_fact_check}.
Directly generated summaries (``w/o SWI'') tend to be more lengthy and verbose, resulting in higher recall scores.
In contrast, SWI-generated summaries exhibit greater accuracy, conciseness, and factual correctness, with fewer hallucinations.
Overall, SWI consistently outperforms the direct generation baseline in terms of F1 score.

\begin{table}[t!]
    \centering
    \scalebox{0.68}{
    \begin{tabular}{ccccccc}
    \toprule
    \midrule
    & & \textbf{Method} & \textbf{GSM8K} & \textbf{GSM8K-P} & \textbf{MATH500} & Avg. \\
    \midrule
    \multirow{2}{*}{CoT} & \ding{192} & w/o SWI & 77.86 & 80.07 & 42.00 & 66.64 \\
    & \ding{193} & \cellcolor{swi_yellow}\multirow{1}{*}{\textbf{SWI}} & \cellcolor{swi_yellow}\textbf{80.21} & \cellcolor{swi_yellow}\textbf{82.55} & \cellcolor{swi_yellow}\textbf{42.80} & \cellcolor{swi_yellow}\textbf{68.52} \\
    \midrule
    \multirow{2}{*}{PS} & \ding{194} & w/o SWI & 72.56 & 75.35 & 40.00 & 62.64 \\
    & \ding{195} & \cellcolor{swi_yellow}\multirow{1}{*}{\textbf{SWI}} & \cellcolor{swi_yellow}\textbf{79.45} & \cellcolor{swi_yellow}\textbf{82.54} & \cellcolor{swi_yellow}\textbf{41.40} & \cellcolor{swi_yellow}\textbf{67.80} \\
    \midrule
    \bottomrule
    \end{tabular}
    }
    \caption{\textbf{LLM Reasoning \& Planning with SWI.} When additional LLM reasoning (CoT) and planning (PS) methods are adopted, the exact matching scores (\%) on multiple \textbf{math} datasets with or without SWI.}
    \label{tab:exp_math_swi_cot_ps}
\end{table}

\begin{table}[t!]
    \centering
    \scalebox{0.58}{
    \begin{tabular}{cccccccc}
    \toprule
    \midrule
    & & \multirow{2}{*}{\textbf{Method}} & \multicolumn{3}{c}{News Article} & \multicolumn{1}{c}{Dialogue} & \multicolumn{1}{c}{Wiki Article} \\
    \cmidrule(lr){4-6} \cmidrule(lr){7-7} \cmidrule(lr){8-8}
    & & & \textbf{CDM} & \textbf{XSum} & \textbf{XL-Sum} & \textbf{DialogSum} & \textbf{WikiLingua} \\
    \midrule
    \multirow{2}{*}{CoT} & \ding{192} & w/o SWI & 23.17 & 11.54 & 11.11 & 15.77 & 14.44 \\
    & \ding{193} & \cellcolor{swi_yellow}\multirow{1}{*}{\textbf{SWI}} & \cellcolor{swi_yellow}\textbf{24.25} & \cellcolor{swi_yellow}\textbf{13.86} & \cellcolor{swi_yellow}\textbf{13.73} & \cellcolor{swi_yellow}\textbf{19.49} & \cellcolor{swi_yellow}\textbf{16.88} \\
    \midrule
    \multirow{2}{*}{PS} & \ding{194} & w/o SWI & 24.12 & 12.21 & 11.91 & 17.92 & 15.86 \\
    & \ding{195} & \cellcolor{swi_yellow}\multirow{1}{*}{\textbf{SWI}} & \cellcolor{swi_yellow}\textbf{24.43} & \cellcolor{swi_yellow}\textbf{12.46} & \cellcolor{swi_yellow}\textbf{12.28} & \cellcolor{swi_yellow}\textbf{18.95} & \cellcolor{swi_yellow}\textbf{16.76} \\
    \midrule
    \bottomrule
    \end{tabular}
    }
    \caption{\textbf{LLM Reasoning \& Planning with SWI.} When additional LLM reasoning (CoT) and planning (PS) methods are adopted, the ROUGE scores (\%) on multiple \textbf{summarization} datasets with or without SWI.}
    \label{tab:exp_sum_swi_cot_ps}
    \vspace{-5pt}
\end{table}

\subsection{Synergy with Other Methods}
\label{subsec:synergy_with_cot_ps}

In recent years, various methods have been proposed to boost the reasoning and planning abilities of LLMs.
Since our SWI method is orthogonal to previous work, it is necessary to compare the performance and study the synergy between SWI and them.
As mentioned in \S~\ref{subsec:exp_setup_baseline}, we adopt representative LLM reasoning method Chain-of-Thought (CoT)~\citep{kojima2022cot_think_step_by_step} and LLM planning method Plan-and-Solve (PS)~\citep{wang2023plan}.

When additional reasoning (CoT) and planning (PS) methods are adopted, Table~\ref{tab:exp_math_swi_cot_ps} and Table~\ref{tab:exp_sum_swi_cot_ps} present the LLM performance on multiple math and summarization tasks, respectively.
Comparing the results in Table~\ref{tab:exp_main_results} and Table~\ref{tab:exp_math_swi_cot_ps}, our SWI method (\ding{193} in Table~\ref{tab:exp_main_results}) beats CoT (\ding{192} in Table~\ref{tab:exp_math_swi_cot_ps}) and PS (\ding{194} in Table~\ref{tab:exp_math_swi_cot_ps}).
Moreover, the combination of CoT+SWI (\ding{193} in Tables \ref{tab:exp_math_swi_cot_ps} and \ref{tab:exp_sum_swi_cot_ps}) boosts the CoT method (i.e., \ding{193}$>$\ding{192}), and the synergy of PS+SWI (\ding{195} in Tables \ref{tab:exp_math_swi_cot_ps} and \ref{tab:exp_sum_swi_cot_ps}) also improves the PS-alone performance (i.e., \ding{195}$>$\ding{194}).
These results verify that SWI works synergistically with existing LLM reasoning \& planning methods.

\subsection{Generalizability to Different LLMs}
\label{subsec:more_llms}

To further validate the generalizability of our SWI method, we evaluate its effect on different sizes and types of LLMs.
Aside from the results of LLaMA3-8B (Table~\ref{tab:exp_main_results}), we present the results on multiple mathematical reasoning tasks using LLaMA3-3B (with 3B parameters) and LLaMA3-8B-R1, which is fine-tuned using reasoning data distilled from DeepSeek R1~\citep{guo2025deepseek_r1}.
As observed in Table~\ref{tab:exp_math_more_llm}, our SWI method brings consistent improvements over the direct generation baseline, verifying the effectiveness of SWI when applied to models of different model sizes and LLM types (i.e., chat and reasoning models).

\begin{table}[t!]
    \centering
    \scalebox{0.58}{
    \begin{tabular}{ccccccc}
    \toprule
    \midrule
    \textbf{Model} & & \textbf{Method} & \textbf{GSM8K} & \textbf{GSM8K-P} & \textbf{MATH500} & Avg. \\
    \midrule
    \multirow{2}{*}{LLaMA3-3B} & \ding{192} & w/o SWI & 45.64 & 46.82 & 27.20 & 39.89 \\
    & \ding{193} & \cellcolor{swi_yellow}\multirow{1}{*}{\textbf{SWI}} & \cellcolor{swi_yellow}\textbf{65.05} & \cellcolor{swi_yellow}\textbf{67.58} & \cellcolor{swi_yellow}\textbf{32.80} & \cellcolor{swi_yellow}\textbf{55.14} \\
    \midrule
    \multirow{2}{*}{LLaMA3-8B-R1} & \ding{194} & w/o SWI & 68.08 & 70.72 & 56.40 & 65.07 \\
    & \ding{195} & \cellcolor{swi_yellow}\multirow{1}{*}{\textbf{SWI}} & \cellcolor{swi_yellow}\textbf{75.44} & \cellcolor{swi_yellow}\textbf{79.24} & \cellcolor{swi_yellow}\textbf{57.00} & \cellcolor{swi_yellow}\textbf{70.56} \\
    \midrule
    \bottomrule
    \end{tabular}
    }
    \caption{\textbf{Generalizability of SWI to different LLMs.} When different sizes and types of LLMs are adopted, the exact matching scores (\%) on multiple mathematical reasoning datasets with or without SWI.}
    \label{tab:exp_math_more_llm}
\end{table}

\begin{table}[t!]
    \centering
    \scalebox{0.6}{
    \begin{tabular}{ccccccc}
    \toprule
    \midrule
    & \multirow{2}{*}{\textbf{Method}} & \multicolumn{3}{c}{News Article} & \multicolumn{1}{c}{Dialogue} & \multicolumn{1}{c}{Wiki Article} \\
    \cmidrule(lr){3-5} \cmidrule(lr){6-6} \cmidrule(lr){7-7}
    & & \textbf{CDM} & \textbf{XSum} & \textbf{XL-Sum} & \textbf{DialogSum} & \textbf{WikiLingua} \\
    \midrule
    \ding{192} & \multirow{1}{*}{w/o SWI} & 23.38 & 11.90 & 11.29 & 16.92 & 15.01 \\
    \midrule
    \rowcolor{swi_yellow}\ding{193} & \multirow{1}{*}{SWI} (V0) & 23.91 & 13.90 & 13.80 & 19.57 & 16.53 \\
    \midrule
    \ding{194} & SWI (V1) & 24.27 & 14.12 & 14.10 & 19.43 & 17.34 \\
    \ding{195} & SWI (V2) & 24.04 & 14.69 & 14.66 & 19.24 & 17.06 \\
    \ding{196} & SWI (V3) & 24.17 & 13.83 & 14.02 & 18.82 & 16.10 \\
    \midrule
    \bottomrule
    \end{tabular}
    }
    \caption{\textbf{Results of Different SWI Prompt Variants.} When different paraphrases of SWI prompts are adopted, the ROUGE scores (\%) on various text summarization datasets with or without Speaking with Intent (SWI).}
    \label{tab:exp_prompt_variant_sum}
    \vspace{-10pt}
\end{table}

\subsection{SWI Prompt Variants}
\label{subsec:prompt_variants}

In this work, we implement SWI in a straightforward prompting way for simplicity and reproducibility.
To demonstrate that SWI works effectively irrespective of specific prompt design, we conduct experiments on different SWI prompt variants.
Specifically, the original SWI prompt (``V0'') is paraphrased into three different versions (details in the appendix) by GPT-4o~\citep{openai2024gpt4o}. Table~\ref{tab:exp_prompt_variant_sum} shows the performance of each SWI variants over multiple summarization tasks.
As illustrated, the proposed SWI method, regardless of its prompt implementation, maintains a consistent advantage over the direct generation baseline, substantiating that SWI is an effective and general framework that brings intent---a key cognitive concept for reasoning and communication---into LLM generation.

\begin{table}[t!]
    \centering
    \scalebox{0.6}{
    \begin{tabular}{cccccccc}
    \toprule
    \midrule
    \multirow{2}{*}{\textbf{Task}} & \multirow{2}{*}{\textbf{Dataset}} & \multicolumn{3}{c}{\textbf{\# Input Tokens}} & \multicolumn{3}{c}{\textbf{\# Output Tokens}} \\
    \cmidrule(lr){3-5} \cmidrule(lr){6-8}
    & & w/o SWI & SWI & $\Delta$ & w/o SWI & SWI & $\Delta$ \\
    \midrule
    \multirow{5}{*}{\textbf{Sum}} & CDM & 920 & 1028 & +108 & 166 & 434 & +161\% \\ 
    & XSum & 542 & 650 & +108 & 135 & 374 & +177\% \\ 
    & XL-Sum & 613 & 721 & +108 & 150 & 405 & +170\% \\ 
    & DialogSum & 263 & 371 & +108 & 77 & 238 & +209\% \\ 
    & WikiLingua & 525 & 633 & +108 & 145 & 386 & +166\% \\ 
    \midrule
    \multirow{3}{*}{\textbf{QA}} & BBH & 203 & 313 & +110 & 96 & 244 & +154\% \\ 
    & MMLU & 193 & 303 & +110 & 63 & 229 & +263\% \\ 
    & MMLU-Pro & 270 & 380 & +110 & 493 & 703 & +43\% \\ 
    \midrule
    \multirow{3}{*}{\textbf{Math}} & GSM8K & 123 & 233 & +110 & 211 & 280 & +33\% \\ 
    & GSM8K-P & 123 & 233 & +110 & 213 & 273 & +28\% \\ 
    & MATH500 & 133 & 243 & +110 & 810 & 922 & +14\% \\ 
    \midrule
    \bottomrule
    \end{tabular}
    }
    \caption{\textbf{Efficiency Study.} The number of input and output tokens with or without SWI.}
    \label{tab:stat_efficiency}
    \vspace{-10pt}
\end{table}

\subsection{Efficiency Study}

To study the efficiency cost of requiring LLMs to speak with intent, we present the number of input and output tokens with or without SWI in Table~\ref{tab:stat_efficiency}.
SWI adds about $110$ extra input tokens in the system prompt (as in Table~\ref{tab:swi_instruction}) for additional instructions, and the extra output tokens are not more than 300 tokens (percentages in Table~\ref{tab:stat_efficiency}) on average compared to direct generation.
Hence, this cost should be traded off against the substantially improved performance, transparency, and interpretability of SWI.

\begin{figure*}[t!]
  \centering
  \includegraphics[width=0.98\textwidth]{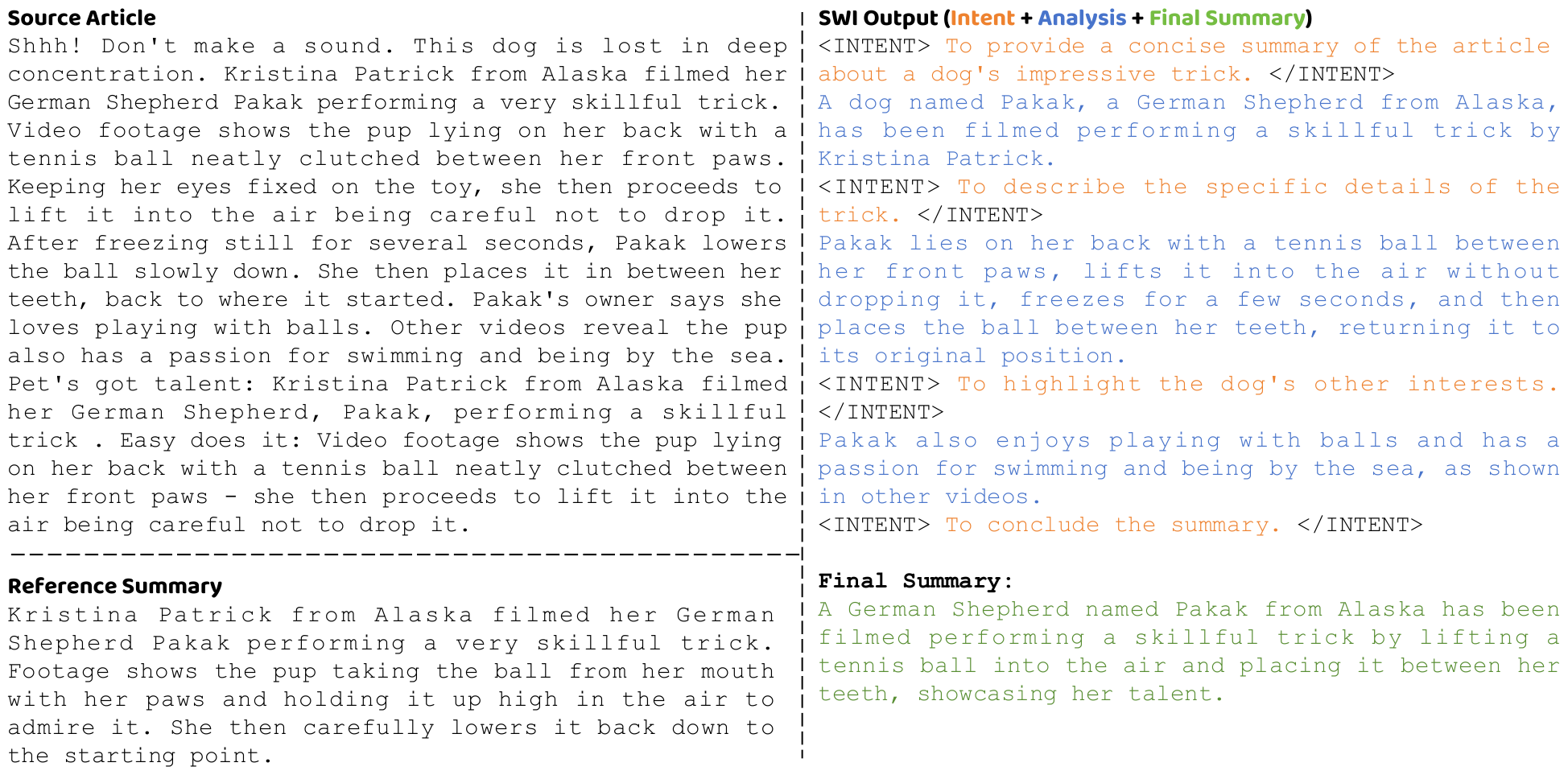}
  \vspace{-3pt}
  \caption{\textbf{Case Study.} The source article, reference summary, and SWI output of a text summarization example.}
  \label{fig:case_study}
  \vspace{-10pt}
\end{figure*}

\subsection{Case Study}
\label{subsec:case_study}

In addition to the math example in Figure~\ref{fig:swi_showcase}, we conduct a case study on the summarization task to provide more insights into the benefits of SWI.
Figure~\ref{fig:case_study} showcases the SWI outputs, where the generated intent is well formulated and articulated, guiding a point-by-point summarization process that leads to a final summary that is accurate, concise, and abstractive, while effectively capturing the key information of the source article.
This probably leads to the high Precision and F1 scores of SWI in Table~\ref{tab:exp_sum_fact_check}, suggesting the validity of SWI in text generation tasks.
Similarly, SWI enables LLM to have progressive planning when solving math problems.
This is a critical ability when dealing with complex problems that require the divide-and-conquer strategy.

\begin{table}[t!]
    \centering
    \scalebox{0.7}{
    \begin{tabular}{ccccc}
    \toprule
    \midrule
    \textbf{Task} & \textbf{Dataset} & \textbf{Total} & \textbf{Unique} & \textbf{Per Ins.} \\
    \midrule
    \multirow{5}{*}{\textbf{Sum}} & CDM & 53,086 & 86 & 4.6 \\
    & XSum & 50,697 & 92 & 4.5 \\
    & XL-Sum & 56,100 & 84 & 4.9 \\
    & DialogSum & 4,739 & 96 & 3.2 \\
    & WikiLingua & 12,463 & 91 & 4.2 \\
    \midrule
    \multirow{3}{*}{\textbf{QA}} & BBH & 7,425 & 49 & 1.3 \\
    & MMLU & 19,380 & 106 & 1.4 \\
    & MMLU-Pro & 21,916 & 122 & 1.8 \\
    \midrule
    \multirow{3}{*}{\textbf{Math}} & GSM8K & 5,237 & 45 & 4.0 \\
    & GSM8K-P & 4,719 & 43 & 3.9 \\
    & MATH500 & 3,619 & 84 & 7.2 \\
    \midrule
    \bottomrule
    \end{tabular}
    }
    \vspace{-5pt}
    \caption{\textbf{Intent Statistics.} The number of total and unique intents (verbs) of each dataset, as well as the average number of intents per instance (``Per Ins.'').}
    \label{tab:stat_intent}
    \vspace{-12pt}
\end{table}

\subsection{Intent Statistics}

To further analyze the pattern and variability of the intents generated by our SWI method, we present the intent statistics across different tasks.
Specifically, we extract and count the verbs in the generated intent statements, which are required to be in the format ``To do something.''.
Table~\ref{tab:stat_intent} shows the number of total verbs (``Total'') and unique verbs (``Unique'') of each dataset, as well as the average number of intents per instance (``Per Ins.'').
We observe that the number of unique intents in the summarization task is larger, indicating that summarizing documents demands a higher intent variability.
In addition, summarization and math tasks generally have more intents per instance than QA, likely due to their longer outputs, as also observed in the efficiency study (Table~\ref{tab:stat_efficiency}).
Among the three math datasets, MATH500 is relatively harder, as the model performance is lower in Table~\ref{tab:exp_main_results}. Thus, solving the problems in MATH500 requires more thinking steps and longer reasoning chains, which is consistent with the observation that its number of unique intent verbs and the average number of intents per instance are larger than GSM8K.

\begin{figure}[t!]
  \centering
  \begin{subfigure}[b]{0.48\linewidth}
    \includegraphics[width=\linewidth]{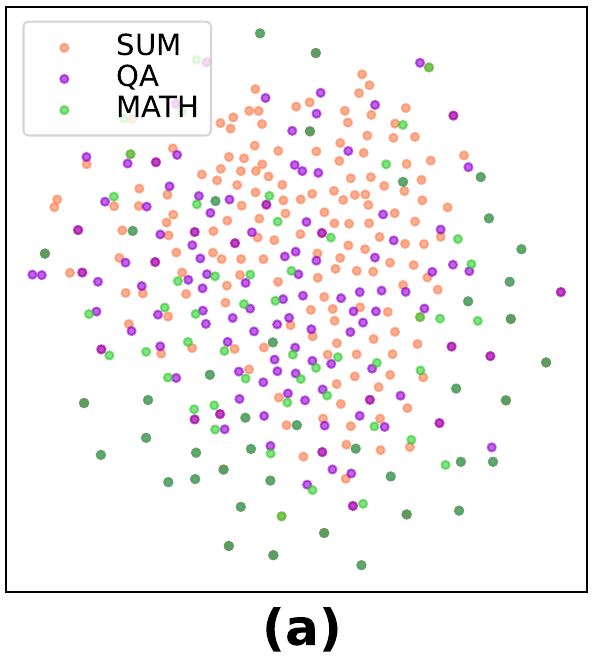}
    \label{fig:intent_distribution_all_tasks}
  \end{subfigure}
  \vspace{-10pt}
  \begin{subfigure}[b]{0.48\linewidth}
    \includegraphics[width=\linewidth]{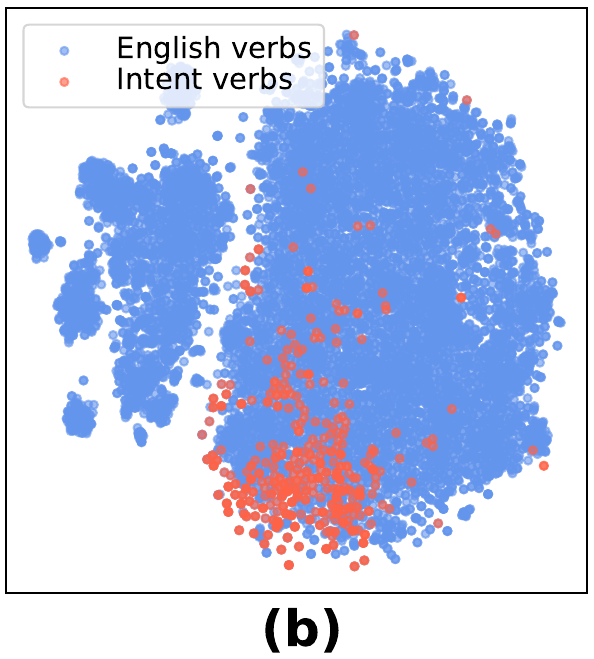}
    \label{fig:intent_distribution_all_vs_eng}
  \end{subfigure}
  \vspace{-10pt}
  \caption{\textbf{The semantic distribution of intents} across different task types and among all English verbs.}
  \label{fig:intent_distribution}
  \vspace{-15pt}
\end{figure}

\begin{figure}[t!]
  \centering
  \begin{subfigure}[b]{0.32\linewidth}
    \includegraphics[width=\linewidth]{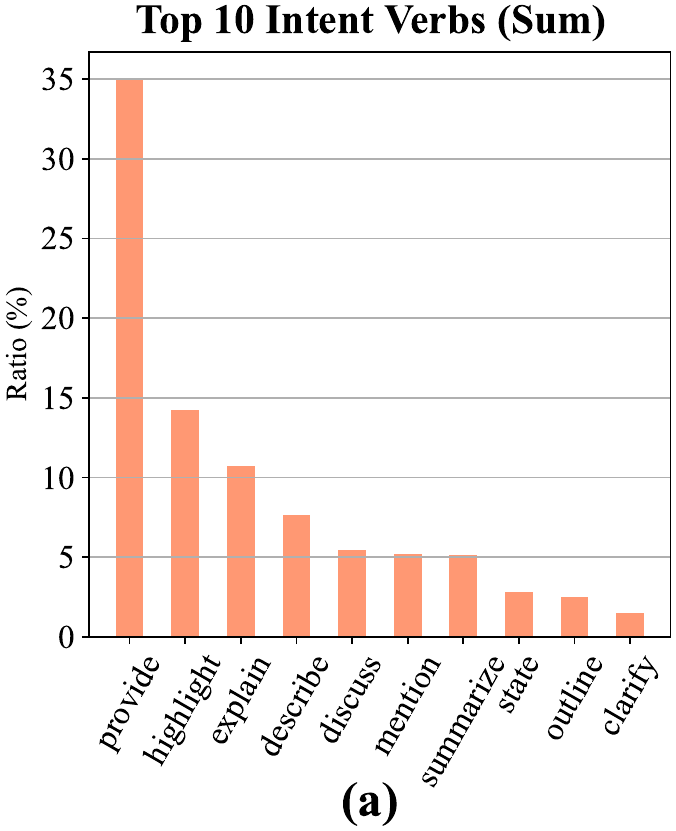}
    \label{fig:intent_stat_sum}
  \end{subfigure}
  \vspace{-10pt}
  \begin{subfigure}[b]{0.32\linewidth}
    \includegraphics[width=\linewidth]{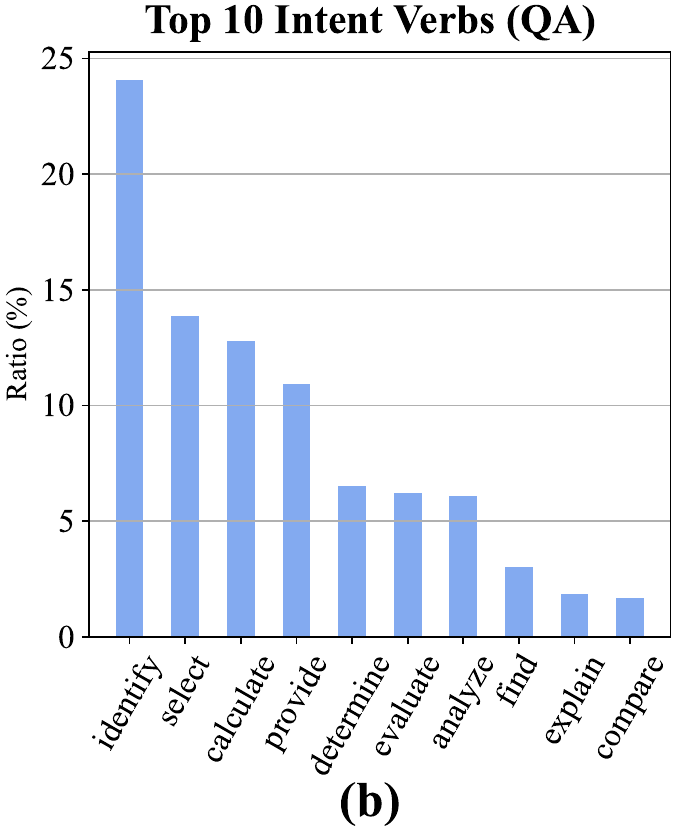}
    \label{fig:intent_stat_qa}
  \end{subfigure}
  \vspace{-10pt}
  \begin{subfigure}[b]{0.32\linewidth}
    \includegraphics[width=\linewidth]{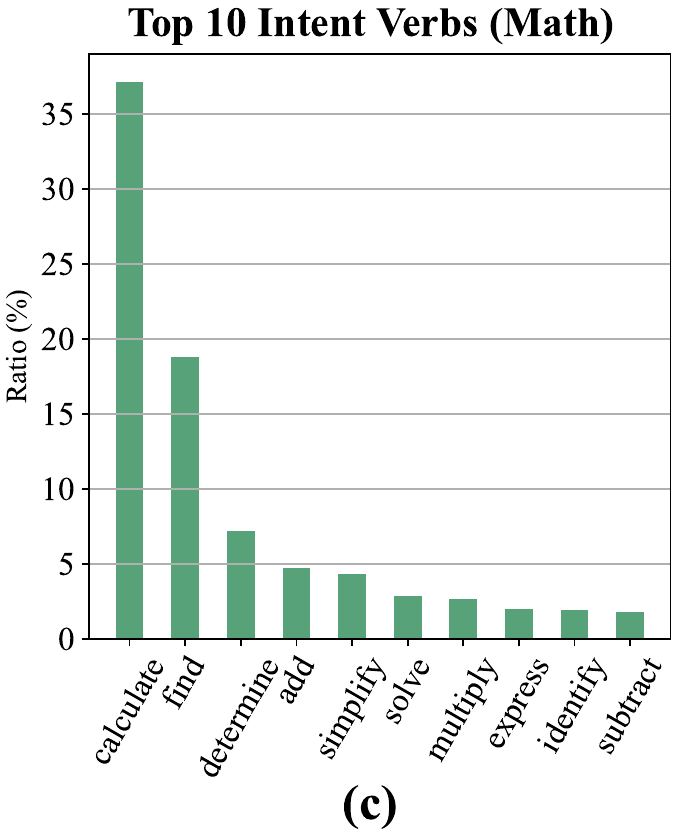}
    \label{fig:intent_stat_math}
  \end{subfigure}
  \caption{\textbf{Top 10 common intent verbs} of (a) summarization, (b) question answering, and (c) math tasks.}
  \label{fig:intent_stat}
  \vspace{-15pt}
\end{figure}

Furthermore, we investigate the intents across different task types and the distribution of intent verbs among all English verbs.
First, we feed the same model that generates the outputs (i.e., LLaMA3-8B) with each verb and extract the last-layer hidden states, which indicate how the generator perceives and utilizes the intents.
Then, t-SNE~\citep{vandermaaten2008tSNE} is applied to visualize the semantic representations of each unique intent verb and all 11,531 English verbs drawn from WordNet~\citep{miller1992wordnet}.
We observe from Figure~\ref{fig:intent_distribution}(a) that all three tasks involve diverse intents, indicating the need for versatile intent skill sets when performing different tasks.
Figure~\ref{fig:intent_distribution}(b) illustrates that intent verbs mainly lie in a certain cluster, showing their specialty from common English verbs.
In addition, Figure~\ref{fig:intent_stat} presents the top 10 common intent verbs in summarization, QA, and math tasks, demonstrating the nuance of intents used in different tasks.

\vspace{-5pt}
\subsection{Intent Quality Evaluation}
\label{subsec:intent_eval}

Although we have shown that SWI boosts performance across a broad range of tasks, verifying the quality of generated intents is also significant.
Thus, we hire human evaluators to assess the quality of generated intent across three criteria: coherence, effectiveness, and interpretability.
Coherence measures how well the intent guides analysis and reasoning, effectiveness evaluates its contribution to problem-solving, and interpretability assesses its role in enhancing user understanding of the generated content.
For each instance, human evaluators are provided with evaluation instructions, task input (e.g., the math problem, question with options, or source article), SWI-generated output, and assessment check boxes. They are then asked to evaluate the following aspects:

\vspace{-5pt}
\begin{itemize}
\setlength\itemsep{0em}
    \item \textbf{Coherence}: \textit{In general, does the analysis align coherently with the intent statements?}
    \item \textbf{Effectiveness}: \textit{Overall, do the intent statements help with the planning and reasoning for performing the task?}
    \item \textbf{Interpretability}: \textit{Do you think providing the intent can help you better understand the reasoning process than not providing it?}
\end{itemize}
\vspace{-5pt}

\begin{table}[t!]
    \centering
    \scalebox{0.55}{
    \begin{tabular}{cccccccc}
    \toprule
    \midrule
    \multirow{2}{*}{\textbf{Task}} & \multirow{2}{*}{\textbf{Dataset}} & \multicolumn{2}{c}{\textbf{Coherence}} & \multicolumn{2}{c}{\textbf{Effectiveness}} & \multicolumn{2}{c}{\textbf{Interpretability}} \\
    &  & Score & Agree & Score & Agree & Score & Agree \\
    \midrule
    \multirow{2}{*}{Summarization} & CDM & \cellcolor{teagreen}2.83 & 80\% & \cellcolor{swi_green_dark}2.77 & 70\% & \cellcolor{teagreen}2.83 & 75\% \\
    & XSum & \cellcolor{swi_green_dark}2.70 & 70\% & \cellcolor{swi_green_dark}2.60 & 65\% & \cellcolor{swi_green_dark}2.57 & 65\% \\
    \midrule
    \multirow{2}{*}{Math Reasoning} & GSM8K & \cellcolor{teagreen}2.90 & 85\% & \cellcolor{teagreen}2.97 & 95\% & \cellcolor{teagreen}2.97 & 95\% \\
    & MATH500 & \cellcolor{teagreen}2.87 & 80\% & \cellcolor{teagreen}2.87 & 80\% & \cellcolor{teagreen}2.83 & 80\% \\
    \midrule
    \multirow{2}{*}{Multi-task QA} & BBH & \cellcolor{swi_green}2.37 & 55\% & \cellcolor{swi_green}2.37 & 50\% & \cellcolor{swi_green}2.33 & 45\% \\
    & MMLU & \cellcolor{swi_green_dark}2.67 & 75\% & \cellcolor{swi_green_dark}2.53 & 55\% & \cellcolor{swi_green}2.37 & 45\% \\
    \midrule
    \bottomrule
    \end{tabular}
    }
    \vspace{-3pt}
    \caption{\textbf{Intent Quality Evaluation by Humans.} The score ranges from 1 (Bad) to 3 (Good).}
    \label{tab:exp_intent_eval}
    \vspace{-15pt}
\end{table}

Evaluation scores range from 1 (Bad), 2 (Fair), to 3 (Good).
Agreement ratios are calculated as follows: 1 if all three evaluators agree, 0.5 if two agree, and 0 if all scores differ.
As shown in Table~\ref{tab:exp_intent_eval}, human evaluation scores for all aspects across datasets exceed 2.3, indicating that the generated intent is generally well-regarded.
Notably, we observe very strong assessment scores (near 3) with substantial agreement (approaching 100\%) for both summarization and math tasks, demonstrating that SWI-generated intent is particularly coherent, effective, and interpretable.

The relatively low (but still fairly good) scores observed in QA tasks may be attributed to the lack of multi-step guidance:
as presented in Table~\ref{tab:stat_intent}, the average number of intents per instance in QA tasks is often 1 or 2, which is much lower than that in summarization and math tasks.
This finding indicates the advantages of multi-round iterative intents, with SWI being able to boost task performance even with a few intents generated.

\section{Related Work}
\label{sec:related_work}

\paragraph{Intent in NLG and LLMs.}
Since the seminal work~\citep{grosz1986attention}, intent has played a critical role in NLG.
In the classical NLG pipeline~\citep{reiter1997nlg,reiter2000nlg}, content determination and document planning are modeled as a process of communicative goals decomposition and ordering, where the resulting planned communicative acts the NLG system wants to achieve are its intentions.
Most approaches followed this framework by implementing a computational model of the Rhetorical Structure Theory (RST)~\citep{mann1988rhetorical}, a discourse theory that explains how parts of a multi-sentential text relate to each other functionally, i.e., how each piece serves a communicative purpose relative to the whole.
For instance, RST has been applied in NLG to many genres, ranging from handling explanation dialogues~\citep{moore1993planning} to generating persuasive evaluative arguments~\citep{carenini2006generating}.
In all these applications, the text was planned before being generated and intentions were explicit.

In contrast, in modern LLM-driven generation, intent is typically implicit. In other words, what we see is only the generated text, with no access to the underlying communicative goals and corresponding intentions.
In this respect, SWI can be seen as the first attempt to make LLM intentionality more explicit, bridging the gap between classic NLG and LLMs, arguably boosting LLMs' controllability and interpretability.

\paragraph{LLM Reasoning.}
Early LLMs were rather poor at reasoning~\citep{openai2018gpt,openai2019gpt2}, and scaling pre-training was shown not to be a feasible solution for improving reasoning~\citep{chu2025post_training}.
Instead, Chain-of-Thought (CoT) prompting~\citep{kojima2022cot_think_step_by_step,wei2022cot} demonstrated that by modifying the prompt, LLMs can elicit a beneficial step-by-step reasoning process at test time without additional training~\citep{li2024cot_transformer,yeo2025demystifying,zhang2025tts_survey}.
Building on CoT, various reasoning techniques have emerged~\citep{xu2025lrm}. Among them, a recent research ARR~\citep{yin2025arr} consistently outperforms CoT on multiple QA tasks, where analyzing the intent of questions is its most effective component.
Different from ARR, we enable LLMs to articulate their intent, using it to guide subsequent analysis and reasoning for improved task performance.

Inspired by the success of CoT and similar prompting techniques, very recent research is increasingly focusing on enhancing LLMs reasoning abilities by explicitly training them for reasoning using reinforcement learning (RL)~\citep{sutton2018rl,shao2024deepseek_math}.
Intriguingly, the success of SWI demonstrated in this paper could spur a similar explosion of research on training LLMs with RL~\citep{wang2025offline_rl,setlur2025rewarding} to better analyze and formulate intentions.

\paragraph{Intent-related Research.}
Intent Detection (ID) and New Intent Discovery (NID)~\citep{kumar2022intent_detection,liang2024conversational_intent,zhang2024intent_detection,tang2024intent_discovery,zhang2024intent_encoder,qian2024user_intention,yin2025midlm}, which classify utterances into known or novel intent categories, are longstanding challenges in natural language understanding~\citep{larson2019intent_detection,casanueva2020intent_detection,zhang2021discovering,weld2022survey_intent_detection}.
Typically, these tasks are approached as classification problems~\citep{wang2024user_intent,yoon2024blendx,zhang2024intent_discovery,sakurai2024evaluating_intention}, where models assign sentences to predefined intent classes.
On the contrary, our SWI method generates intent as free-form text rather than fixed categories, enhancing flexibility and fluency.
SWI naturally integrates intent statements as planning into the reasoning process, providing contextual guidance for subsequent analysis.

\section{Conclusion}
\label{sec:conclusion}

In this work, we introduce Speaking with Intent (SWI) in LLMs, where the generated intent (as high-level planning) guides subsequent analysis, improving the generation and reasoning abilities.
Extensive experiments across text summarization, multi-task QA, and mathematical reasoning benchmarks consistently show the benefits of speaking with explicit intent over the direct generation baseline.
In text summarization, SWI produces summaries that are more accurate, concise, and factually reliable, with fewer hallucinations.
In addition, SWI outperforms existing LLM reasoning and planing methods and works synergistically with them.
Further analysis substantiates the generalizability of SWI when applied to different settings.
Moreover, human evaluations solidify the coherence, effectiveness, and interpretability of LLM-generated intent.
Overall, this study opens a new avenue for enhancing LLM generation and reasoning abilities.

\section*{Ethics \& Impact Statement}

This work does not raise ethical issues, and we would like to mention the impact of SWI.
As intent is a fundamental aspect of natural language processing, empowering, eliciting, and enhancing the intent understanding and generation abilities can potentially drive AI systems (including multimodal models) to the next level.
Moreover, Speaking with Intent can also be applied to various domains beyond NLP, such as healthcare, law, and finance. These applications are cost-sensitive, so explicitly showing the intent of AI models will help with the transparency and interpretability of critical decision-making.

\section*{Acknowledgments}

Nous remercions le Conseil de recherches en sciences naturelles et en g\'{e}nie du Canada (CRSNG) de son soutien.

We acknowledge the support of the Natural Sciences and Engineering Research Council of Canada (NSERC), Vector Institute for AI, and Accelerate Foundation Models Research Program Award from Microsoft.
This research was supported in part by the computational resources and services provided by Advanced Research Computing at the University of British Columbia and the Digital Research Alliance of Canada (alliancecan.ca).
We would also like to thank Vered Shwartz for constructive feedback on conducting evaluations.

\clearpage
\bibliography{swi}

\clearpage
\appendix

\section{Experiment Details}
\label{app:exp_details}

\subsection{Dataset Details}
\label{app:dataset_details}

All datasets used in this work are loaded from Hugging Face datasets. Table~\ref{tab:dataset_sources} lists the URL link of each dataset.
Please note that the URLs may be subject to change by the dataset providers.

\begin{table}[ht]
    \centering
    \scalebox{0.8}{
    \begin{tabular}{clcc}
    \toprule
    \midrule
    \textbf{Task} & \multicolumn{1}{l}{\textbf{Dataset}} & \textbf{Source} \\
    \midrule
    \multirow{5}{*}{\makecell[c]{Sum}} & CDM~\citep{hermann2015cnn_dailymail_1} & \href{https://huggingface.co/datasets/abisee/cnn_dailymail}{Link} \\
    & XSum~\citep{narayan2018xsum} & \href{https://huggingface.co/datasets/EdinburghNLP/xsum}{Link} \\
    & XL-Sum~\citep{hasan2021xl_sum} & \href{https://huggingface.co/datasets/GEM/xlsum}{Link} \\
    & DialogSum~\citep{chen2021dialogsum} & \href{https://huggingface.co/datasets/knkarthick/dialogsum}{Link} \\
    & WikiLingua~\citep{ladhak2020wikilingua} & \href{https://huggingface.co/datasets/GEM/wiki_lingua}{Link} \\
    \midrule
    \multirow{3}{*}{\makecell[c]{QA}} & BBH~\citep{suzgun2023bbh} & \href{https://huggingface.co/datasets/lukaemon/bbh}{Link} \\
    & MMLU~\citep{hendrycks2021mmlu} & \href{https://huggingface.co/datasets/hails/mmlu_no_train}{Link} \\
    & MMLU-Pro~\citep{wang2024mmlu_pro} & \href{https://huggingface.co/datasets/TIGER-Lab/MMLU-Pro}{Link} \\
    \midrule
    \multirow{3}{*}{\makecell[c]{Math}} & GSM8K~\citep{cobbe2021gsm8k} & \href{https://huggingface.co/datasets/openai/gsm8k}{Link} \\
    & GSM8K-P~\citep{vendrow2025gsm8k_platinum} & \href{https://huggingface.co/datasets/madrylab/gsm8k-platinum}{Link} \\
    & MATH500~\citep{lightman2024math500} & \href{https://huggingface.co/datasets/HuggingFaceH4/MATH-500}{Link} \\
    \midrule
    \bottomrule
    \end{tabular}
    }
    \caption{\textbf{Dataset Sources.}}
    \label{tab:dataset_sources}
\end{table}

\subsection{Model Details}
\label{app:model_details}

As mentioned in \S~\ref{subsec:exp_setup_model}, we mainly employ LLaMA3-8B-Instruct~\citep{meta2024llama3}, an instruction-following LLM with 8 billion model parameters, for most experiments. In generalizability experiments (\S~\ref{subsec:more_llms}), we also explore LLMs of different sizes and types.
Table~\ref{tab:model_sources} presents the URL link of each model and tokenizer provided by Hugging Face Transformers~\citep{wolf2020transformers}.

\begin{table}[ht]
    \centering
    \scalebox{0.8}{
    \begin{tabular}{lc}
    \toprule
    \midrule
    \multicolumn{1}{l}{\textbf{Model}} & \textbf{Source} \\
    \midrule
    LLaMA3-3B~\citep{meta2024llama3} & \href{https://huggingface.co/meta-llama/Llama-3.2-3B-Instruct}{Link} \\
    LLaMA3-8B~\citep{meta2024llama3} & \href{https://huggingface.co/meta-llama/Llama-3.1-8B-Instruct}{Link} \\
    LLaMA3-8B-R1~\citep{guo2025deepseek_r1} & \href{https://huggingface.co/deepseek-ai/DeepSeek-R1-Distill-Llama-8B}{Link} \\
    \midrule
    \bottomrule
    \end{tabular}
    }
    \caption{\textbf{Model Sources.}}
    \label{tab:model_sources}
\end{table}

\subsection{SWI Prompt Variants}
\label{app:prompt_variants}

As mentioned in \S~\ref{subsec:method_swi}, we implement SWI in a straightforward prompting way for simplicity and reproducibility, i.e., we require LLMs to speak with intent (SWI) by presenting detailed instructions in the system prompts and restating the SWI requirement in the user prompt.
In addition, we paraphrase the SWI prompt into three different versions (\S~\ref{subsec:prompt_variants}) to demonstrate that our SWI method maintains effectiveness irrespective of the specific prompt formulation.
Here, we present the prompt variants in Table~\ref{tab:prompt_variants}.

\begin{table*}[ht]
    \centering
    \scalebox{0.8}{
    \begin{tabular}{cl}
    \toprule
    \midrule
    \multicolumn{1}{l}{\textbf{Type \& Version}} & \textbf{Prompt Text} \\
    \midrule
    \makecell[c]{System Prompt \\ V0 (default)} & \makecell[l]{You are a helpful assistant who speaks with intent. You are good at summarizing \\ documents and the summary must start with "Final Summary:" \\ During generation, follow all the requirements below: \\
    1. Always explicitly state your own intent before speaking each sentence. \\
    2. Each intent statement should explain the sentence that follows. \\
    3. Your intent must start with the "<INTENT>" tag and end with the "</INTENT>" tag. \\ The content within the intent tags must begin with \\ "To" followed by a verb, such as "To accomplish a task." \\
    4. At last, clearly and concisely give your final summary starting with "Final Summary:"} \\
    \midrule
    \makecell[c]{System Prompt \\ V1} & \makecell[l]{You are a purposeful assistant skilled in document summarization who speaks with intent. \\ Your final response must begin with "Final Summary:" \\ While generating responses, adhere strictly to these instructions: \\
    1. Before every sentence, clearly state your intent using an explanation. \\
    2. Each intention should directly clarify the sentence that follows. \\
    3. Use the tags "<INTENT>" and "</INTENT>" to wrap each intent statement. \\ Each statement inside the intent tags must begin with "To" and a verb, \\ for example, "To describe the process." \\
    4. Conclude with a clear and concise final summary that begins with "Final Summary:"} \\
    \midrule
    \makecell[c]{System Prompt \\ V2} & \makecell[l]{You are a helpful assistant who is skilled in text summarization and always communicates \\ with deliberate intent. Ensure your final output starts with "Final Summary:" \\
    Comply with the following instructions during your response: \\
    1. Begin each sentence with a description of your intent. \\
    2. The intent must directly relate to and explain the sentence that comes after it. \\
    3. Surround each intent with the tags "<INTENT>" and "</INTENT>". Each intent \\ statement enclosed by the tags should start with the word "To" and an action verb, \\
    like "To explain the reasoning." \\
    4. Finish with a succinct summary, introduced by "Final Summary:"} \\
    \midrule
    \makecell[c]{System Prompt \\ V3} & \makecell[l]{You are a precise and helpful assistant proficient in text summarization, who always \\ speaks with deliberate intent. Your final response must begin with "Final Summary:" \\ While producing your response, follow these guidelines: \\
    1. Before each sentence, declare your intent explicitly. \\
    2. Ensure each intent explains the sentence that immediately follows. \\
    3. Wrap every intent declaration with "<INTENT>" and "</INTENT>" tags. \\ Make sure that every intent statement within the tags begins with "To" and an action verb, \\
    for example, "To justify the choice." \\
    4. Conclude your response with a clearly stated final summary prefaced by "Final Summary:"} \\
    \midrule
    \makecell[c]{User Prompt \\ All Versions} & \makecell[l]{Speak with intent and summarize the following document. \\ \texttt{\{\{article\}\}}} \\
    \midrule
    \bottomrule
    \end{tabular}
    }
    \caption{\textbf{SWI Prompt Variants.}}
    \label{tab:prompt_variants}
\end{table*}

\section{Human Evaluation Details}
\label{app:human_eval}

\paragraph{Participant Requirements.}
We hire human evaluators from the cloud-sourcing platform CloudResearch to conduct human evaluation on the quality of the generated intent: coherence, effectiveness, and interpretability.
To ensure the annotation quality, we apply several requirements to select qualified human evaluators, as shown in Table~\ref{tab:human_eval_requirements}.

\begin{table}[ht]
    \centering
    \scalebox{0.6}{
    \begin{tabular}{cl}
    \toprule
    \midrule
    \textbf{Type} & \multicolumn{1}{c}{\textbf{Requirements}} \\
    \midrule
    Native Language & English \\
    \midrule
    Country of Residence & Australia, Canada, Ireland, New Zealand, UK, US \\
    \midrule
    Education & Undergraduate student, Graduate student \\
    \midrule
    \multirow{2}{*}{Reputation} & Approved Projects Count: $\geq$ 1,000 \\
    & Approval Rating: $\geq$ 90\% \\
    \midrule
    \bottomrule
    \end{tabular}
    }
    \caption{\textbf{The requirements for human evaluators.}}
    \label{tab:human_eval_requirements}
\end{table}

\paragraph{Evaluation Tasks.}
For each task category, we select two datasets: CDM~\citep{hermann2015cnn_dailymail_1,see2017cnn_dailymail_2} and XSum~\citep{narayan2018xsum} for text summarization, BBH~\citep{suzgun2023bbh} and MMLU~\citep{hendrycks2021mmlu} for multi-task multiple-choice QA, and GSM8K~\citep{cobbe2021gsm8k} and MATH500~\citep{lightman2024math500} for mathematical reasoning.
We randomly sample 12 instances per dataset and divide them into two batches of six.
Each batch includes a dummy instance with deliberately reversed intents to ensure evaluators are actively engaged rather than randomly selecting responses.
Evaluator submissions are accepted or rejected based on completion time and performance on the dummy instance.

For each instance, human evaluators are provided with evaluation instructions, task input (e.g., the math problem, question with options, or source article), SWI-generated output, and assessment check boxes. They are then asked to evaluate the following aspects:
\begin{itemize}
\setlength\itemsep{0em}
    \item \textbf{Coherence}: \textit{In general, does the analysis align coherently with the intent statements?}
    \item \textbf{Effectiveness}: \textit{Overall, do the intent statements help with the planning and reasoning for performing the task?}
    \item \textbf{Interpretability}: \textit{Do you think providing the intent can help you better understand the reasoning process than not providing it?}
\end{itemize}

Each batch is assessed by three different human evaluators, with each person uniquely assigned to only one batch.
Evaluation scores range from 1 (Bad), 2 (Fair), to 3 (Good).
Agreement ratios are calculated as follows: 1 if all three evaluators agree, 0.5 if two agree, and 0 if all scores differ.

\paragraph{Human Evaluation Quality.}
As mentioned above, we decided to accept or reject the evaluator's submission based on the task completion time and the results on the dummy instance that is deliberately modified to have a lower coherence.
As a result, about 60\% of the evaluators still rated the dummy instance as good coherence, meaning they failed the dummy test and potentially did not fully focus on the evaluation process, which poses a general caveat to the quality of cloud-sourcing annotations.
Overall, we rejected about 10\% of submissions that both failed the dummy test and took an unreasonably short time to complete the annotation.
After rejecting them, we hired other evaluators until the intent quality evaluation was finished.

\paragraph{Human Evaluation Cost.}
The pay rate for each human evaluator is US\$10 per hour, completing a batch of 6 instances takes an evaluator 10-15 minutes on average, and the total cost of the intent quality evaluation is about US\$120.

\end{document}